\documentclass[letterpaper, 10 pt, conference]{ieeeconf} 

\IEEEoverridecommandlockouts 
\usepackage{amsfonts}
\usepackage{amsmath}
\usepackage{float}
\usepackage{color}
\usepackage{tikz}
\usepackage{subcaption}
\usepackage{multicol}
\usepackage{float}
\usepackage[]{algorithm2e}
\usepackage{upgreek}
\usepackage{epsf}
\usepackage{bm}
\usepackage{amsmath} 
\usepackage{amssymb}
 \usepackage{epsfig}
 \usepackage{graphicx}  %Required
 \usepackage{psfrag}

\usepackage{graphicx,psfrag}	
\usepackage[latin1]{inputenc}

\title{\LARGE \bf
Robot Navigation in Dynamic Environments using Acceleration Obstacles
}

\author{Asher Stern and Zvi Shiller$^{1}$
\thanks{$^{1}$ Mechanical Engineering and Mechatronics, Ariel University, Israel.  
{\tt\small shiller@ariel.ac.il}}%
}

\begin{document}
\maketitle
\pagestyle{empty}

\begin{abstract}
This paper addresses the issue of motion planning in dynamic environments by extending the concept of Velocity Obstacle (\textit{VO}) \cite{fiorini-7} and Nonlinear Velocity Obstacle (\textit{NLVO}) \cite{Shiller-2001} to Acceleration Obstacle (\textit{AO}) and Nonlinear Acceleration Obstacle (\textit{NAO}).  Similarly to \textit{VO} and \textit{NLVO}, the \textit{AO}and \textit{NAO} represent the set of colliding constant \textit{accelerations}  of the maneuvering robot with obstacles moving along linear and nonlinear trajectories, respectively.  Contrary to prior works, we derive analytically the \textit{exact} boundaries  of \textit{AO}  and \textit{NAO}.  

To enhance an intuitive understanding of these representations, we first derive the \textit{AO} in several steps:  first extending the \textit{VO} to the Basic Acceleration Obstacle (\textit{BAO}) that consists of the set of constant accelerations of the robot that would collide with an obstacle moving at \textit{constant} accelerations, while assuming zero initial velocities of the robot and obstacle. This is then extended to the \textit{AO} while assuming arbitrary initial velocities of the robot and obstacle.  And finally, we derive the  \textit{NAO} that in addition to the prior assumptions, accounts for obstacles moving along \textit{arbitrary} trajectories.   
  
The introduction of \textit{NAO} allows the generation of safe avoidance maneuvers that directly account for the robot's second-order dynamics, with acceleration as its the control input.  The \textit{AO} and \textit{NAO} are demonstrated in several examples of selecting avoidance maneuvers in challenging road traffic.  It is shown that the use of \textit{NAO} drastically reduces the adjustment rate of the maneuvering robot's acceleration while moving in complex road traffic scenarios.  The presented approach enables reactive and efficient navigation for multiple robots, with potential application for autonomous vehicles operating in complex dynamic environments.  
\end{abstract}

\section{Introduction}

Dynamic environments represent an important and growing segment of modern
automation with applications as diverse as, ground,
aerial and marine autonomous vehicles, air and sea traffic control,
automated wheelchairs and even virtual animation and virtual reality games.
Common to these applications is the need for a decision system able to quickly select maneuvers that avoid potential collisions with static and moving obstacles, while moving towards a specified goal.  The challenge of such a decision system is its ability to avoid collision with any number of static and moving obstacles and reach the goal while considering robot dynamics and the trajectories, known or estimated, of the surrounding moving obstacles.
This is a serious challenge since the connectivity of the configuration space in dynamic environments, and hence the goal's reachability may change during motion. 
The main objective of the planner in dynamic environments is therefore to ensure the survival of the robot while attempting to reach the goal.  

An effective approach to avoid collisions in dynamic environments is the use of the Velocity Obsacle (VO) \cite{fiorini-7} that maps obstacles, static or dynamic, to the velocity space of the maneuvering robot.  The  Velocity Obstacle (VO), represents the set of colliding velocities between the robot and an individual obstacle.  Selecting a velocity outside the VO of all obstacles ensures collision-free motion while the obstacle is moving at a constant velocity.  The VO was extended to the Nonlinear Velocity Obstacle (NLVO), which accounts for arbitrary known or predicted trajectories of the obstacle \cite{Shiller-2001}. It allows much fewer velocity adjustments than the linear version \cite{fiorini-7} when the obstacle is moving along curved trajectories.

Another variant of the VO is the Reciprocal VO (RVO) \cite{reciprocal2008} \cite{bareiss2015generalized}. It assumes multi-robot avoidance where each robot is expected to contribute to the avoidance effort.  Geometrically, the RVO is a scaled version of the original VO so that each robot makes only a partial effort to avoid the other obstacle (by avoiding a smaller VO), letting the other robot reciprocate by sharing the mutual avoidance maneuver.  It was claimed that this avoids oscillations that were attributed to the original VO.    

In this paper, we address the obstacle avoidance problem in the acceleration domain by extending the VO to AO (Acceleration Obstacle) and the NLVO to NAO (Nonlinear Acceleration Obstacle).  This is motivated by the fact that a robot moving in a dynamic environment is a dynamic and not a kinematic system.  The simplest model for such a system is of second order that is driven by acceleration that can be arbitrarily selected subject to the robot's acceleration constraints.    

The Acceleration Obstacle, AO, in analogy to the Velocity Obstacle, VO, consists of the constant {\it accelerations} that would cause collisions between a robot and a moving obstacle.  Unlike the VO, the geometric shape of the AO depends on the initial velocities of the robot and the obstacle.    

The AO was earlier addressed in \cite{van2011reciprocal}, and more recently in \cite{AO2022}. 
Despite being conceived in  \cite{van2011reciprocal},  the AO was not used then for the reason that accelerations tend to change frequently and are therefore difficult to observe.  They proposed instead the Acceleration Velocity Obstacle, AVO, which is similar to the VO, except that it accounts for the transition from the current to the target velocity using a proportional feedback law on the acceleration.  Our experience shows that the acceleration applied by the moving obstacle, short or long, are crucial in selecting the robot's proper avoidance maneuver (it is often sufficient for a short acceleration to divert the obstacle away from a collision course).  

The  AO was rigorously introduced in \cite{AO2022} in the context of navigation in human crowds. The AO is derived for a robot and obstacle with an initial relative location and velocity, and a constant relative acceleration.  The AO is constructed as a union of disks, each expressing the constant relative acceleration that would cause collision between the robot and the obstacle at a specific time. 
While the union of the temporal AO(t) defines the exact AO, it does not directly define the AO's boundary, for which \cite{AO2022} offers a linear approximation.  A comprehensive review of current literature on motion planning using the Velocity Obstacle paradigm cen bee seen in  \cite{vesentini2024survey}. 
We are not aware of other works that explicitly address AO in the context of motion planning.          
 
 \section{This paper}
In this paper, we focus on a simple and intuitive derivation of the Acceleration Obstacle, AO, and the Nonlinear Acceleration Obstacle, NAO, in analogy to the VO \cite{fiorini-7} and the NLVO \cite{Shiller-2001}.  
The ability to account for arbitrary (nonlinear) trajectories greatly improves the efficiency of the avoidance procedure when obstacles (vehicles) are moving along observed or anticipated nonlinear trajectories, such as during overtaking \cite{gabriel2022}, roundabouts, and turns.  A direct consideration of such trajectories often allows the avoidance of multiple obstacles with a single velocity or acceleration maneuver, as is later demonstrated in this paper.  In contrast, using AO or AVO to avoid such obstacles 
would require frequent adjustments of the respective avoidance maneuvers.   

We begin with an intuitive extension of the VO, 
starting with zero initial velocities, for which the AO is a simple cone, which we call the Basic Acceleration Obstacle, BAO. We then add initial velocities of the robot and obstacle that cause the AO to warp. Finally, we shift the AO by the constant obstacle acceleration to obtain the absolute representation of the AO. This procedure yields directly the exact boundary of AO.  

We continue with the first introduction of the Nonlinear Acceleration Obstacle, NAO, which consists of the constant robot accelerations that would cause a collision with an obstacle that is moving along an arbitrary (nonlinear) trajectory. Here too, the NAO is defined by its exact boundaries.  

 \vspace{5pt}
\noindent \textbf{Main Contributions of this Paper} 
\begin{enumerate}
     \item Introducing the Nonlinear Acceleration Obstacle \textit{NAO}  that accounts for obstacles moving along arbitrary known trajectories
     \item Introducing a simple graphical visualization of the \textit{AO} and \textit{NAO} 
   \item Offering an analytical computation of the boundaries of \textit{AO} and \textit{NAO}
\end{enumerate}

\section{The Basic Acceleration Obstacle, BAO}
We first construct the simplest form of the Acceleration Obstacle, which we call the Basic Acceleration Obstacle, BAO. It represents 
the set of constant accelerations at a given time, that would cause collisions between a robot and an obstacle (static or moving), assuming zero initial velocities of robot and obstacles.  assuming zero initial velocities of robot and obstacles.  The assumption of zero initial velocities makes the construction of the $BAO$  resemble the construction of the original Velocity Obstacle, VO  \cite{fiorini-7}. 

\begin{figure}%[ht]
\centerline{\resizebox{7cm}{!}{\includegraphics{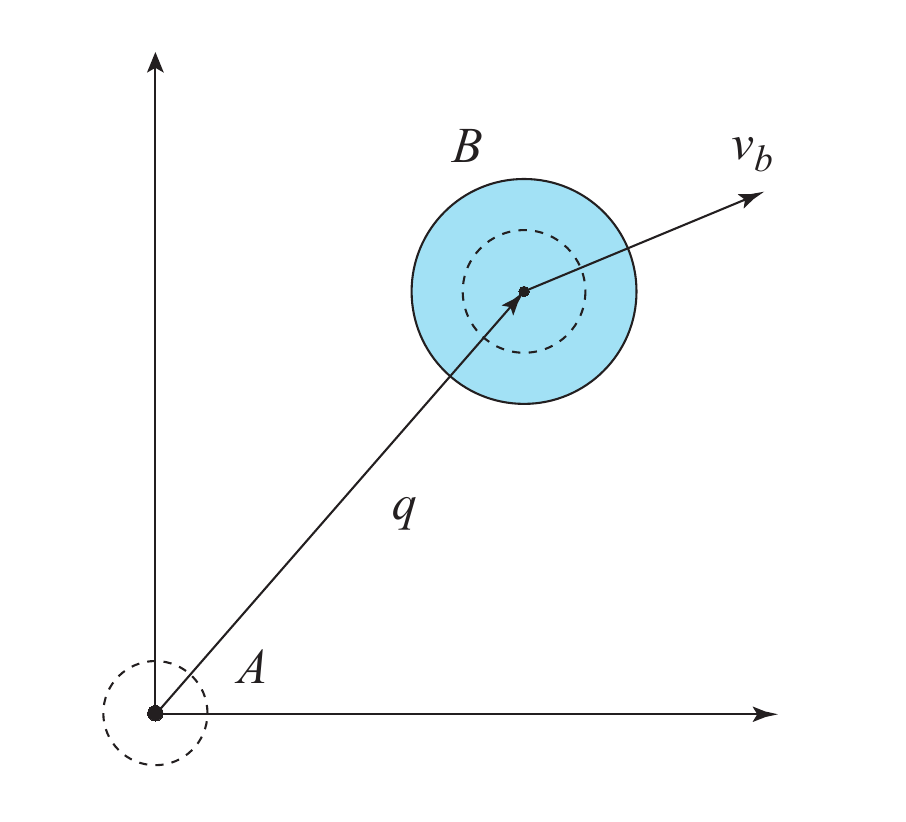}}}
   \caption{A point robot $A$ and a moving obstacle $B$}
   \label{scenario}
 \end{figure}

The geometry of this set 
can be easily described in the configuration space of the robot and obstacles.    
The robot and obstacles can be of general shapes, however, to reduce the dimensionality of the problem, we assume planar circular robots and obstacles.  Growing the obstacle by the
radius of the robot transforms the problem into a point robot avoiding
circular obstacles in the plane as shown in Fig.~\ref{scenario}. 

We denote $A$ as a point robot, located at the origin of an inertial frame; 
$B$ denotes the set of points defining the geometry of an
obstacle, enlarged by the radius of the robot $A$, and $q \in \mathbb{R}^2$, denotes the position of the center of the obstacle in the inertial frame, as shown in Fig.~\ref{scenario}.  
 
The construction of the $BAO$ is demonstrated for the scenario shown in
Fig.~\ref{scenario}, where, at time $t_0$, obstacle
$B$ is moving at a constant acceleration  $a_B$.  
 translating object.    

The $BAO$ at time $t_0$ is constructed by first generating  
the   {\it Relative Acceleration Obstacle} ($RAO$).   
We define 
a {\it ray}  $a/b$, $a,b \in \mathbb{R}^2$, consisting of the half line that originates at $a$ and passes through $b$. 

The $RAO$ is 
defined as the union of all rays originating from $A$ and passing through
$\partial B$, the boundary of $B$ at $(t_0)$: 

\begin{equation} 
RAO = \cup A/b,   b \in \partial B.  
\end{equation}

The set $RAO\subset \mathbb{R}^2$ is the set of all  accelerations of $A$ relative
to $B$, $a_{a/b}$, %\not=0$, 
that would result in collision at some time $t \in (0,\infty)$,
assuming that the obstacle stays on its current course at its current
acceleration.  
Relative accelerations outside of $RAO$ would ensure avoidance of $B$ at all times $t \in (0,\infty)$;  accelerations on the boundary of $RAO$ would result in $A$ grazing $B$.  

\begin{figure}[ht]
 \vspace{.1cm}
\centerline{\resizebox{7cm}{!}{\includegraphics{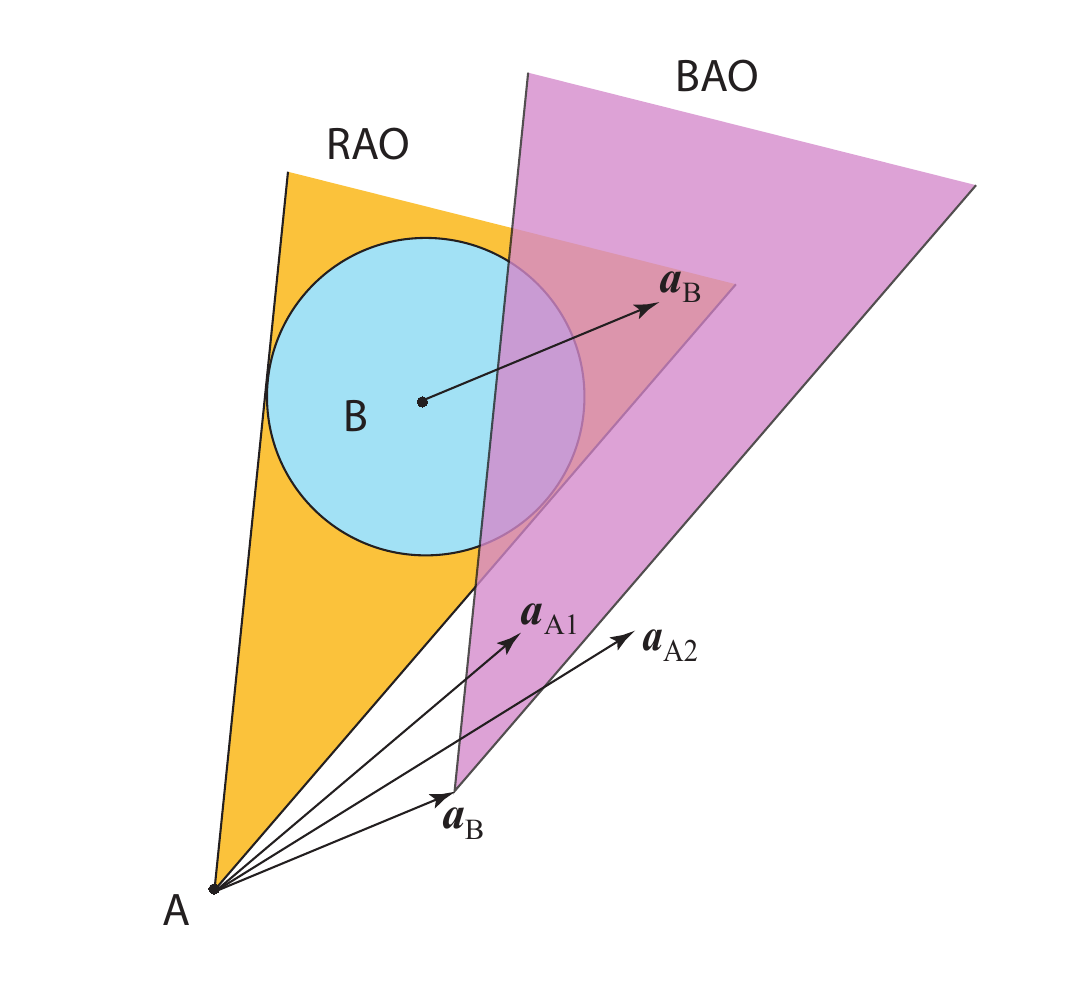}}}
\caption{The Basic Acceleration Obstacle $BAO$}
\label{bao}
\end{figure}

Translating $RAO$ by $A_B$
produces the \emph{Basic Acceleration Obstacle}, 
$BAO\subset \mathbb{R}^2$: 
\begin{equation}
BAO = \boldsymbol{a}_B \oplus RAO.  
\end{equation} 
where $\oplus$ denotes the Minkowski sum.  Thus, $BAO$ represents a set of {\it absolute} accelerations of $A$, $\boldsymbol{a}_a$, that would result in collision at some time $t \in (0,\infty)$. 
In
Fig,~\ref{bao}, $\boldsymbol{a}_{A1}$ is a colliding velocity,
whereas $\boldsymbol{a}_{A2}$ is not.

\vspace{5pt}
\noindent {\bf Definition 1}: The Basic Acceleration Obstacle \\
Consider at time $t_0$ a point robot $A$,
located at the origin of an inertial frame, and an obstacle $B$ centered at $c(t_0)$
and moving at a constant acceleration $a_B$.  The Basic Acceleration obstacle, $BAO$, consists of the
set of all constant accelerations of $A$ at time $t=t_0$ that would collide with $ B$ at any time $t>t_0$:
\begin{equation} 
\label{vo} \
BAO = \{a_A|A(t)\cap B(t) \ne 0\};  t=(t_0,\infty).  
\end{equation} 
 
     It is important to note that the simple cone shape applies only to cases with no initial velocities of the robot and the obstacle. Otherwise, the cone is warped, as is discussed next.   

 \section{Acceleration Obstacle (AO)}
We proceed to account  for nonzero initial velocities $\boldsymbol{v}_{A}$ and $\boldsymbol{v}_{B}$.  This leads to the formation of the {\it Acceleration Obstacle}    $AO\subset \mathbb{R}^2$, which is defined by its boundary. 

\subsection{The exact boundary of $AO$ } 
\begin{figure}[ht]
\vspace{.1cm}
\centerline{\resizebox{7cm}{!}{\includegraphics{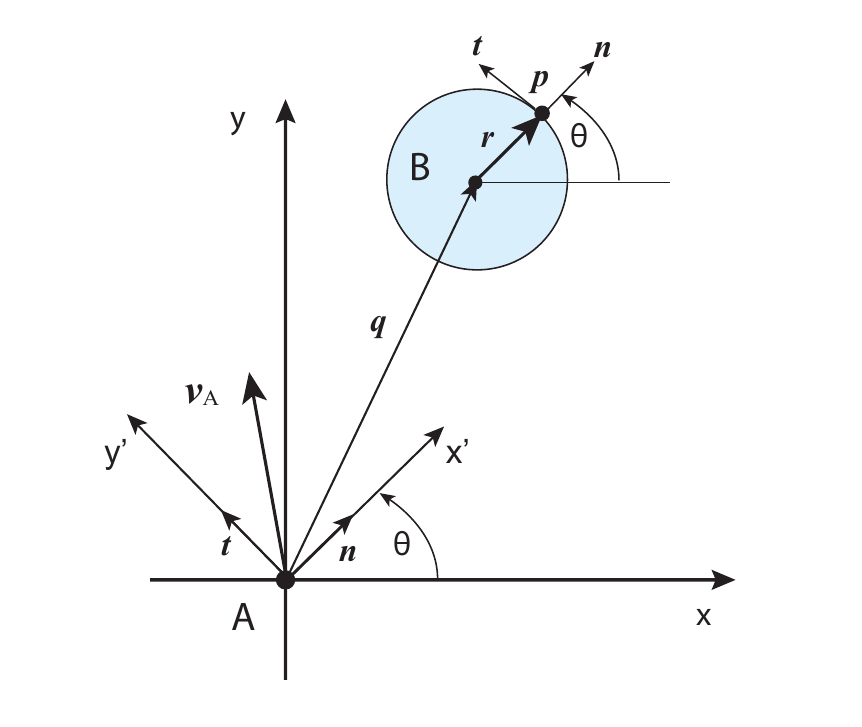}}}
   \caption{A rotated frame along the normal to the obstacle boundary at point $\boldsymbol{p}$}
   \label{AO_p_frame}
 \end{figure}

Consider a point robot $A$ and a circular obstacle $B$ of radius $r$.  For simplicity, we first assume a static obstacle $B$, and $A$ to be moving at an initial velocity $\boldsymbol{v}_{A}$, as shown in Fig.  \ref{AO_p_frame}.  
We wish to compute the set of constant accelerations $\boldsymbol{a}_{A}$ that would cause $A$ to \textit{graze} $B$ along its boundary $\partial B$.  
 
Consider point $ \boldsymbol{p} \in \partial B$, 
$\boldsymbol{p}=\boldsymbol{q}+
r \boldsymbol{n}(\theta)$, where $\boldsymbol{n}(\theta)=e^{i\theta}$ is the normal to $\partial B$ at $\boldsymbol{p}$, and $\boldsymbol{t}(\theta) = i e^{i\theta} $ is the tangent to  $\partial B$ at $\boldsymbol{p} \in \partial B$, as shown in Fig. \ref{AO_p_frame}.  

We first project the vectors $\boldsymbol{v}_A$ and
 $\boldsymbol{p}$ to a coordinate  frame parallel to the unit vectors $\boldsymbol{n}$ and $\boldsymbol{t}$:
\begin{eqnarray}  
p_n &=&  \boldsymbol{p}  \cdot \boldsymbol{n} \label{projections2}\\
p_t &=&  \boldsymbol{p}  \cdot \boldsymbol{t}=\boldsymbol{q}  \cdot \boldsymbol{t} =q_t \nonumber\\
v_{n} &=&  \boldsymbol{v}_{A}  \cdot \boldsymbol{n} \nonumber\\
v_{t}&=&  \boldsymbol{v}_{A}  \cdot \boldsymbol{t}\nonumber
\label{projections}
 \end{eqnarray}
  
We now solve the  following two problems:  

\noindent \textbf{Problem 1:} Find the constant scalar acceleration $a_{n}$ that drives a point mass along $x'$ from $A$ to $\boldsymbol{p}$ between the  boundary conditions:
\begin{eqnarray} 
x'(0)=0&;& x'(t_p)=p_n\;;\;\; x,p_n\in \mathbb{R} \\
\dot{x'}(0) = {v}_{n}&;& \dot{x'}(t_p) = 0; \nonumber
 \end{eqnarray}  
with $t_p>0$, subject to 
\begin{equation} 
\ddot{x'}=a_n = constant
\end{equation}  

\noindent \textbf{Problem 2:}
\noindent Find the constant scalar acceleration $a_{t}$ that drives a point mass along $y'$ from $A$ to $\boldsymbol{p}$ between the  boundary conditions:
\begin{eqnarray} 
y'(0)=0&;& y'(t_p)=p_t\;;\;\; \\
\dot{y'}(0) = {v}_{t}&;& \dot{y'} (t_p) = free; \nonumber
 \end{eqnarray}  
with $t_p>0$, subject to 
\begin{equation} 
\ddot{y'}=a_t = constant
\end{equation}  

Focusing on Problem 1, the constant acceleration $a_{n}$ that satisfies the boundary condition $\dot{x'}(t_p) = 0$   satisfies the integral: 
\begin{equation} 
\int_{v_{n}}^{0}\dot{x}'_nd\dot{x}'=\int_{0}^{p_n}a_{n}dx'
\label{intan}
\end{equation} 
Solving \eqref{intan} for $a_{n}$ yields:
\begin{equation} 
a_{n}=-\frac{v_{n}^2}{2p_n}.
\label{anpn}
\end{equation} 
The time $t_p$  to reach $p_n$ is:
\begin{equation} 
t_p=-\frac{v_{n}}{a_{n}}.
\label{zman}
\end{equation} 
From Problem 2, the equation of motion in the tangent direction $t$ is:
\begin{equation}  
v_{t}t_p+\frac{1}{2}a_tt_p^2=q_t.
\label{tdirection}
\end{equation}  
Substituting \eqref{zman} in \eqref{tdirection} and solving for $a_t$ yields a quadratic equation in $a_n$:
\begin{equation} 
a_t=2\frac{q_t }{v_{n}^2}a_{n}^2 +2\frac{v_{t}}{v_{n}}a_{n}.
\label{att}
\end{equation} 
Equation \eqref{att} describes mathematically the condition  
that $A$ grazes $\partial B$ at $\boldsymbol{p}$. 

\begin{figure}[ht]
\vspace{.1cm}
\centerline{\resizebox{7cm}{!}{\includegraphics{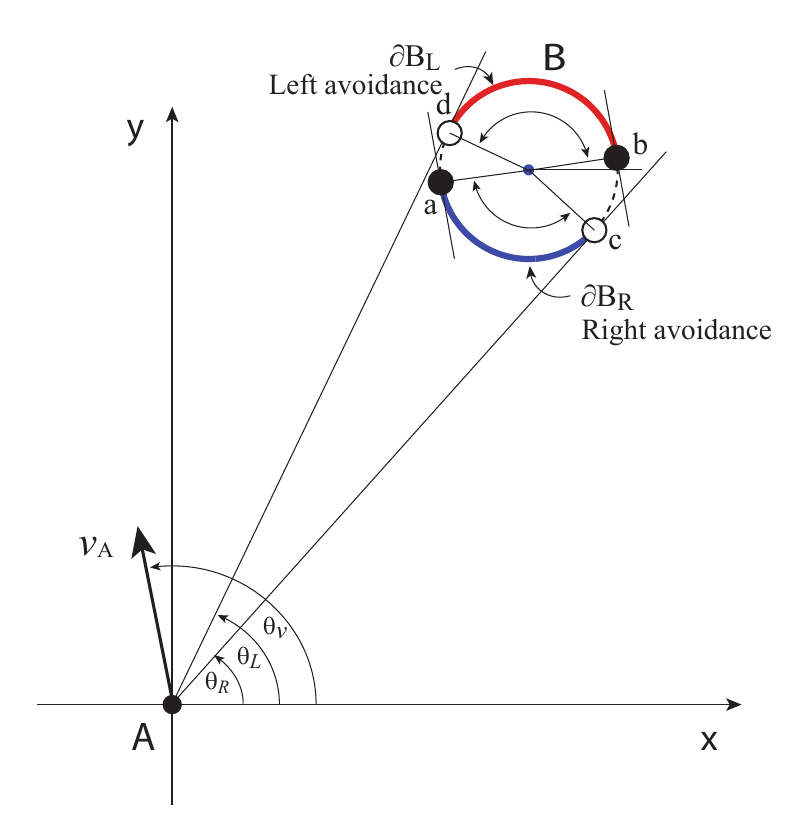}}}
   \caption{Potential grazing points on a static obstacle $B$ for $A$ moving at an  initial velocity $\boldsymbol{v}_A$ and a constant acceleration $\boldsymbol{a}_A$. }
   \label{AO_limits2}
 \end{figure}

Substituting Eq. \eqref{anpn}  into \eqref{zman}  yields the time $t_p$ it would take $A$ to reach  $\boldsymbol{p}$, expressed in terms of the boundary conditions $\boldsymbol{p}$
and $\boldsymbol{v}_A$:    
\begin{equation} 
t_p= \frac{2p_{n}}{v_{n}} >0.
\label{zman2}
\end{equation} 
It follows that   both $p_{n}$ and $v_{n}$ should be of the same sign: 
\begin{equation}  
p_n v_{n} \ge 0. 
\label{segments}
\end{equation} 
Points were either $p_{n}$ or $v_{n}$  changes sign represent boundary points along $\partial B$ that define grazable and nongrazable segments of $\partial B$. Hence, at points where $p_n$ crosses zero, i.e. points $c,d$ in Fig. \ref{AO_limits2},     $t_p\rightarrow 0$.  
Similarly, at points $a,b$,  where $v_n$ crosses zero,  $t_p\rightarrow \infty$.
The segments $a-d$ and $b-c$ are therefore not reachable tangentially. 

The acceleration vector $(a_n,a_t)$ in Equation \eqref{att}  is expressed in the rotated $x',y'$ frame.  
Multiplying $(a_n,a_t)$ by the rotation matrix from frame ($x',y'$) to frame ($x,y$), yields the absolute acceleration $\boldsymbol{a}_A$ in the $x,y$ frame:
\begin{equation} 
\boldsymbol{a}_A(\theta)=R(\theta)(a_n,a_t)^T,
\label{aa}
\end{equation} 
where $R(\theta)\in SO(2)$:  
\begin{equation} 
R(\theta) = [\boldsymbol{n}^T \boldsymbol{t}^T]=\begin{bmatrix}cos(\theta)&-sin(\theta)\\ sin(\theta)&cos(\theta)\end{bmatrix}.
\label{rotation}
\end{equation} 

\begin{figure}[ht]
\vspace{.1cm}
\centerline{\resizebox{9cm}{!}{\includegraphics{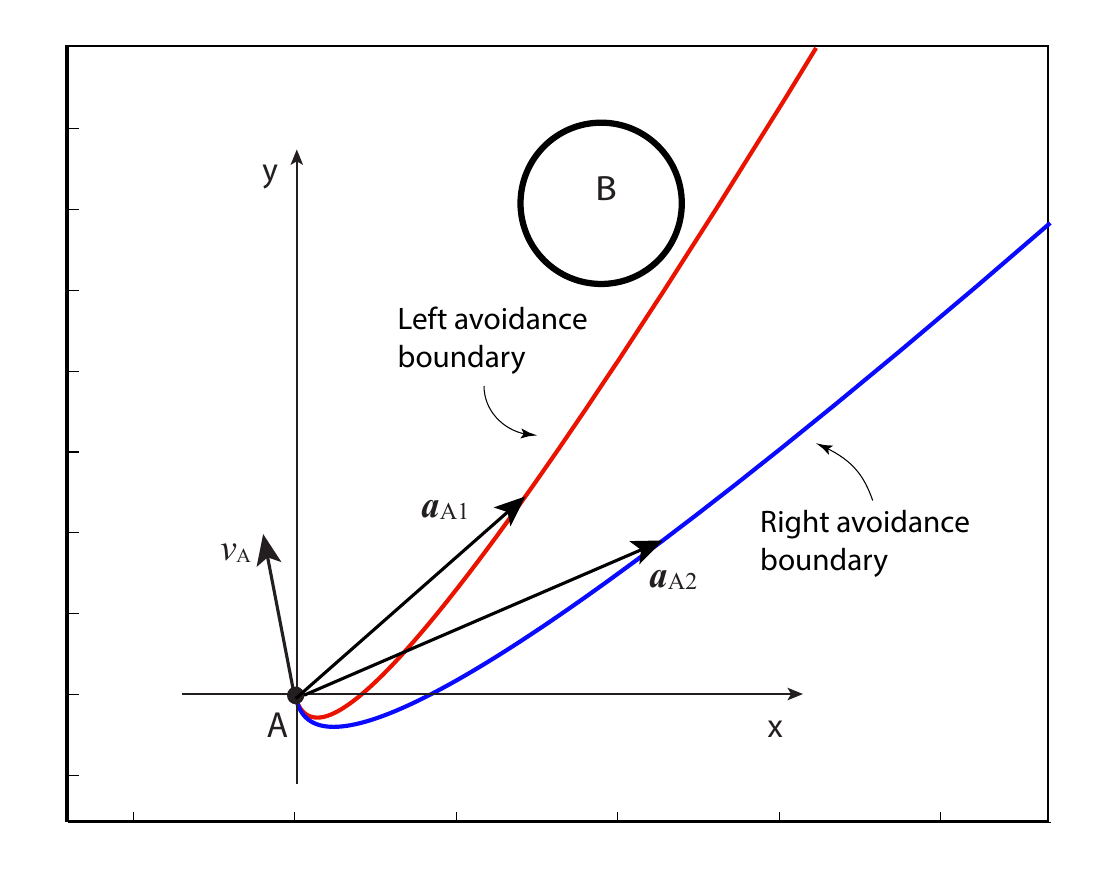}}}
   \caption{The exact boundary of $AO$ for the case shown in Fig. \ref{AO_limits2}.}
   \label{AO_boundaryone}
 \end{figure}

The acceleration $\boldsymbol{a}_A$  \eqref{aa} represents the constant absolute accelerations of $A$ that would result in $A$ grazing $\partial B$. It thus forms  the boundary 

of the Acceleration Obstacle AO shown in Fig. \ref{AO_boundaryone}.  
At points $a,c \in \partial B$,  
$t_p \rightarrow \infty$ 
and $\boldsymbol{a}_A \rightarrow 0 $; 
 the tangent to $\partial B$ at those points  is parallel to $\boldsymbol{v}_A$, as is proven in the Appendix.  At points $b,d \in \partial B$,  
$t_p \rightarrow 0$ and 
$\boldsymbol{a}_A \rightarrow \infty$.  Those points are the tangency points between the cone that originates at $A$  and is tangent to $B$.  This cone coincides with the $RAO$ shown earlier in Fig. \ref{bao}.  Note that the initial velocity $v_A$ has no effect on the trajectory of $A$ when the acceleration $a_A$ approaches infinity.    
Referring to Fig. \ref{AO_limits2}, the arcs $\partial B_R$ and $\partial B_R$ are  defined by their end points $a,c$ and $b,d$ respectively:
\begin{eqnarray} 
\partial B_R = \{\boldsymbol{p}| \boldsymbol{p}=\boldsymbol{q}+re^{i\theta}, \theta_v+\pi/2 <\theta < \theta_R-\pi/2\} \\
\partial B_L = \{\boldsymbol{p}| \boldsymbol{p}=\boldsymbol{q}+re^{i\theta}, \theta_v-\pi/2 <\theta < \theta_L+\pi/2\}  
 \end{eqnarray}
Each boundary arc generates a continuous boundary of $\partial AO$.
We can now formally define the boundaries of $AO$:
 \vspace{5pt}
\noindent \textbf{Definition 1:} Boundary of $AO$
\begin{eqnarray} 
\partial AO_R = \{\boldsymbol{a}_A(\theta)|  \theta \in \partial B_R\} \\
\partial AO_L = \{\boldsymbol{a}_A(\theta)|  \theta \in \partial B_L\}  
 \end{eqnarray}

Fig. \ref{AO_boundaryone} shows the left and right boundaries of the $AO$ generated for a static obstacle $B$ and an initial velocity of $A$, $\boldsymbol{v}_A$.  The boundaries form a warped cone,  originating at $A$.  The cone is warped due to the initial velocity of $A$ and the slope at $A$ of both boundaries coincides with  $\boldsymbol{v}_A$. Also shown in \ref{AO_boundaryone} are two constant accelerations of $A$  that would result in $B$ grazing $A$.  
 
\begin{figure}[ht]
\vspace{.1cm}
\centerline{\resizebox{8cm}{!}{\includegraphics{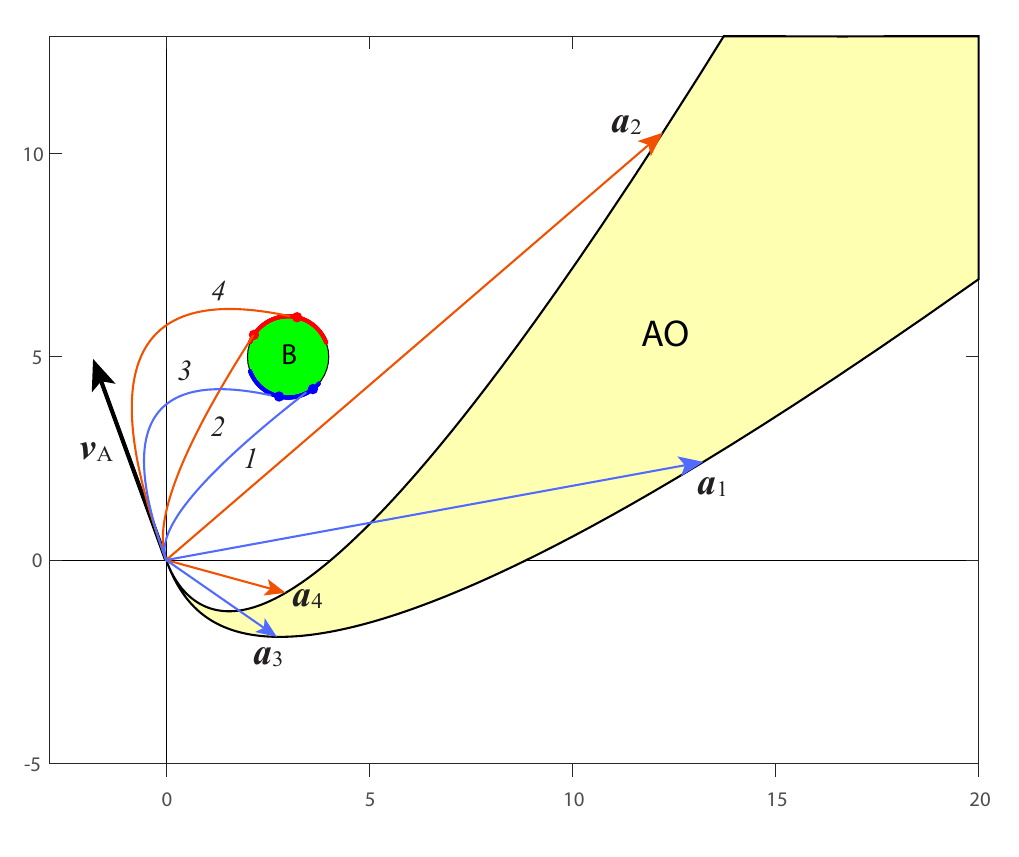}}}
   \caption{Trajectories for selected points on the boundary of AO for a given initial velocity $v_{A}$.  All trajectories are tangent to the boundary of $B$. Trajectories 1 to 4 correspond to accelerations $a_1$ to $a_4$, respectively}
   \label{trajectories}
 \end{figure}

Fig. \ref{trajectories} shows several trajectories generated for accelerations selected along the right and left boundaries of the $AO$ shown in Fig. \ref{AO_boundaryone}. 

\subsection{$AO$ of an obstacle moving at a constant velocity }
 
To account for  an obstacle that is moving at a constant velocity $\boldsymbol{v}_B$, we simply replace the robot velocity $\boldsymbol{v}_A$ with the relative velocity $\boldsymbol{v}_{A/B}$:
\begin{equation} 
\boldsymbol{v}_{A/B} = \boldsymbol{v}_A-\boldsymbol{v}_B.
\end{equation} 
The $AO$ terminates at the origin as in the case of a static obstacle, but the slope of $AO$ at the origin coincides now with the relative velocity $\boldsymbol{v}_{A/B}$.

\subsection{$AO$ of an obstacle moving at a constant acceleration }

\begin{figure}[ht]
\vspace{.1cm}
\centerline{\resizebox{10cm}{!}{\includegraphics{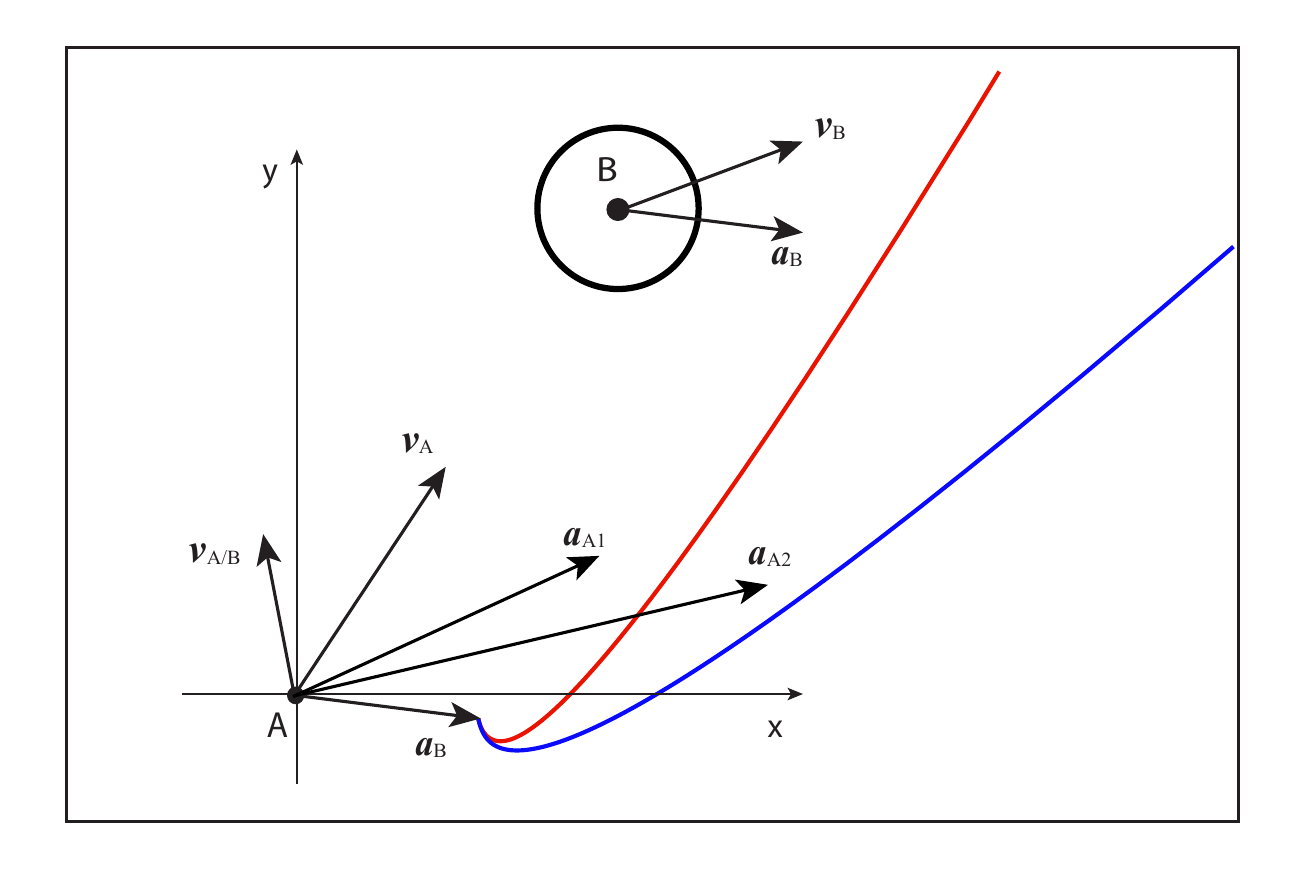}}}
   \caption{$AO$ for an obstacle $B$ moving at an initial velocity $\boldsymbol{v}_B$ and a constant acceleration $\boldsymbol{a}_B$. RAO generated for $v_{A/B}$ and moved by $a_B$}
   \label{AOstar}
 \end{figure}

To account for an obstacle that is moving at a constant acceleration $\boldsymbol{a}_B$, 
we shift  $AO$ by $\boldsymbol{a}_B$ as shown in Fig. \ref{AOstar}, similarly to the shift of $BAO$ as was shown in Fig. \ref{bao}:
\begin{eqnarray} 
\boldsymbol{a}_A =  \boldsymbol{a}_B +\boldsymbol{a}_{A/B}\\
AO=\boldsymbol{a}_B \oplus AO.
 \end{eqnarray}

\begin{figure}[ht]
\vspace{.1cm}
  \centerline{\resizebox{7cm}{!}{\includegraphics{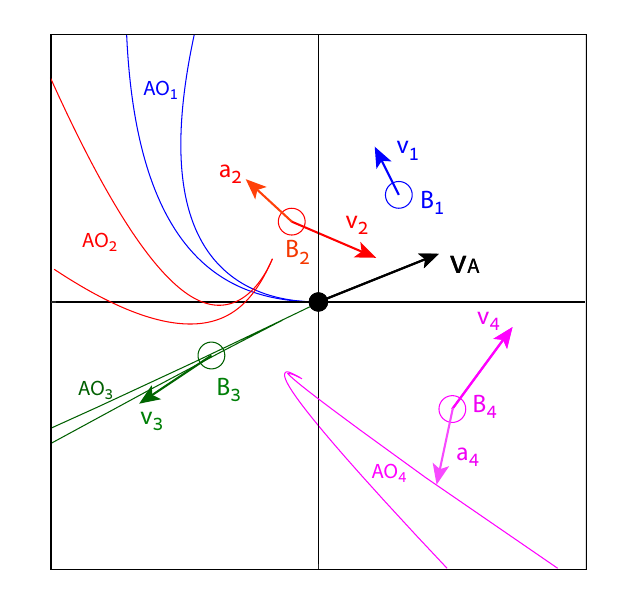}}}
   \caption{The  $AO$ boundaries of several obstacles moving at their respective constant velocities and accelerations with $A$ moving at the initial velocity $\boldsymbol{v}_A$. %All $AO$'s terminate at the origin since they are no.
   }
   \label{fig:AOfull2}
 \end{figure} 

Fig. \ref{fig:AOfull2} shows $AO$'s of four obstacles, each moving at some constant velocity and acceleration, with $A$ moving at an initial velocity $\boldsymbol{v}_{A}$.  Note that the $AO$ of obstacles $B_1$ and $B_3$ with zero acceleration terminates at the origin, whereas obstacles $B_2$ and $B_4$ that accelerate at $a_2$ and $a_4$ are shifted accordingly.    
 
% **************
\section{Nonlinear Acceleration Obstacle (NAO)}
We now address the case of an obstacle moving along an arbitrary trajectory.  This is an extension of the nonlinear Velocity Obstacle (NLVO) \cite{Shiller-2001} developed previously in the velocity space.  Unlike the NLVO, which was constructed as a union of temporal Velocity Obstacles, NLVO(t), the NAO is constructed by computing its {\it exact} boundary for a given trajectory traveled by the obstacle up to a given time horizon.  The advantage of $NAO$ over  $AO$ is obvious, as it requires much fewer acceleration adjustments in cases where the obstacle is moving along a curved trajectory that is either known or observed.   

Consider obstacle $B$, with its center $q$,  following trajectory $c(t)$, and a point robot A that at time $t_0$ is moving at an initial velocity $v_A$, as shown in Fig. \ref{fig:naop}.  We wish to identify the constant accelerations of A at time $t_0$ that would cause collisions with $B$ at any time $t \in (t_0,t_h]$, where $t_h$ is the time horizon until which $c(t)$ is known, observed or estimated. 

The $NAO$  consists of all accelerations of $A$ at $t_0$ that would
result in a collision with the obstacle at any time $t>t_0$.  Selecting a
{\it single} acceleration, $\boldsymbol{a}_A$, at time $t=t_0$ outside of $NAO$ would thus guarantee collision avoidance  at all times:  
\begin{equation} 
(\boldsymbol{v}_A(t_0) t+\frac{1}{2}\boldsymbol{a}_A(t_0)t^2) \cap (\boldsymbol{c}(t)\oplus B)\neq 0
\ ; \ \forall t \in (t_0,t_h]. 
\end{equation}    
It is convenient to define $NAO$ by its boundaries, representing accelerations that would result in $A$ grazing $B$. 
\subsection{The Exact Boundary of $NAO$ }
\begin{figure}%[h]
\vspace{.1cm}
\centerline{\resizebox{7cm}{!}{\includegraphics{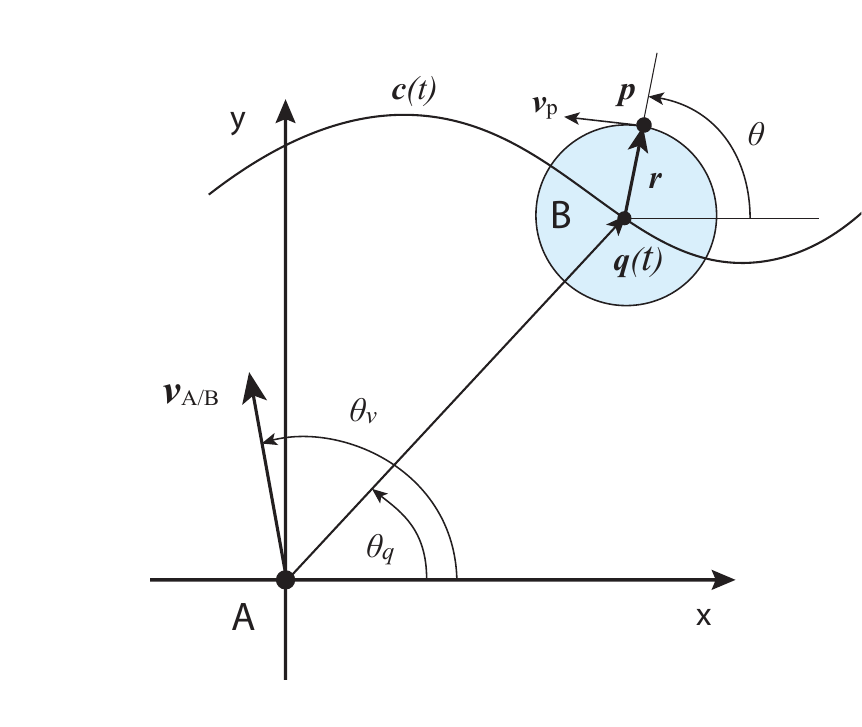}}}
  \caption{Construction of $NAO$.  }
   \label{fig:naop}
 \end{figure}

Consider obstacle $B$ that is moving along trajectory $\boldsymbol{c}(t)$, as shown in Fig. \ref{fig:naop}.  
The boundary of $NAO$ consists of the constant accelerations $\boldsymbol{a}_A$ of $A$ that would cause $A$ to graze $B$ while it moves along $\boldsymbol{c}(t) $ for $t> t_0$.   

We wish to compute the constant acceleration $a_A$ \eqref{ai}, given the initial velocity $v_A$,  that at time $t$ would reach tangentially some point $p\in \partial B$.  At that time, the center of $B$, which coincides with $c(t)$,  is moving at the velocity $\dot c(t)$,  as shown in Fig. \ref{fig:naop}.

Let express the vectors  $v_A, a_A, v_p, v_r, q, r, p$ using complex numbers, referring angles and vectors to Fig. \ref{fig:naop}: 
\begin{eqnarray} 
\boldsymbol{v}_A &=& v_A  e^{i\theta_v}
\label{ai} \\
\boldsymbol{a}_A&=&a_A  e^{i\theta_a} \nonumber
 \\
\boldsymbol{q}&=&qe^{i\theta_q}\nonumber\\
\boldsymbol{r}&=&re^{i\theta}\nonumber\\
\boldsymbol{v}_p&=&v_pe^{i(\theta+\pi/2)}
\nonumber\\
\boldsymbol{v}_r&=&v_re^{i\theta}
\nonumber\\
\boldsymbol{p}&=&qe^{i\theta_q}+re^{i\theta}\nonumber
 \end{eqnarray}
We solve this problem by freezing $B$ at $c(t)$ and subtracting its velocity from the velocity of A: $v_{A/B}=v_A-\dot c(t)$.       

% *******************************

We wish to compute the constant acceleration $a_A$ \eqref{ai}, given the initial velocity $v_A$, that would reach point $p\in \partial B$ tangentially at time $t$,  as was shown in Fig. \ref{fig:naop}. 
\begin{figure}%[h]
\vspace{.1cm}
\centerline{\resizebox{7cm}{!}{\includegraphics{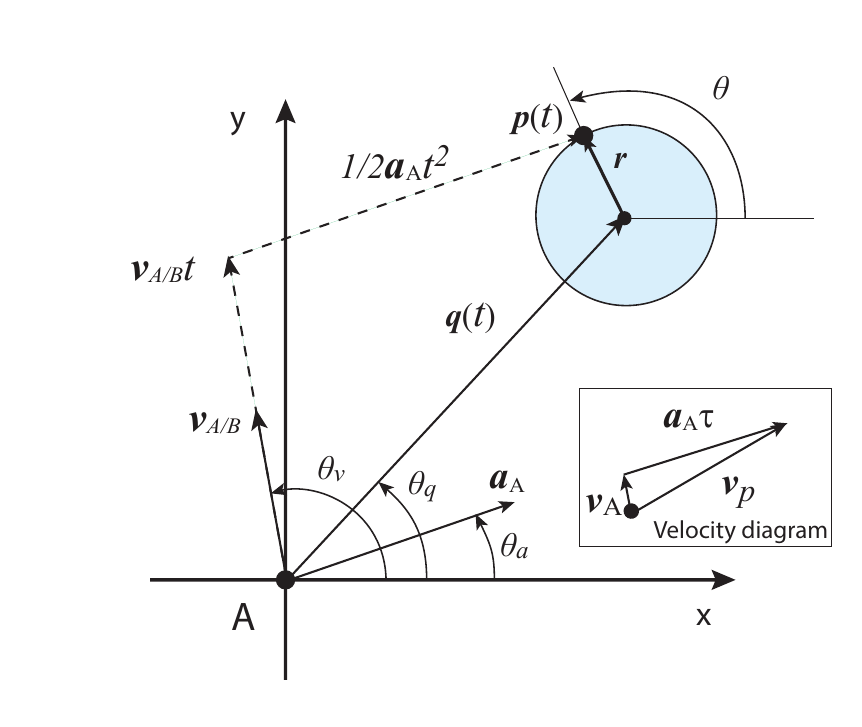}}}
  \caption{Computing the boundary of $NAO$.  }
   \label{fig:naopT}
 \end{figure} 
Referring to Fig. \ref{fig:naopT}, we first equate two paths that reach the point $p(t)$ from $A$:
 \begin{equation} 
\boldsymbol{p}(t)=q(t)e^{i\theta_q}+re^{i\theta(t)}=
v_{A/B}(t)t e^{i\theta_v}+\frac{t^2}{2}a_A(t)  e^{i\theta_a}
\label{47}
 \end{equation} 
 % 
% t is fixed aA, vA, vB are fixed 
For the velocity at $p$, $v_p$, to be tangent to $\partial B$, implies:   
 \begin{equation} 
v_p(t)e^{i(\theta(t)+\pi/2)}=v_{A/B}(t)e^{i\theta_v}+a_A(t)t e^{i\theta_a(t)}
\label{48}
\end{equation} 
Solving \eqref{48} for $a_A$ and substituting into \eqref{47}: 
 \begin{eqnarray} 
p(t) &=& q(t)e^{i\theta_q}+re^{i\theta(t)}\\&=&v_At e^{i\theta_v}+\frac{t}{2}(v_pe^{i\left(\theta(t)+90\right)}-v_Ae^{i\theta_v} ) \nonumber
\label{49}
  \end{eqnarray} 
 Rearranging \eqref{49}:
 \begin{equation} 
q(ct)e^{i\theta_q}-\frac{t}{2}v_Ae^{i\theta_v}=\frac{t}{2}v_pie^{i\theta(t)}-re^{i\theta(t)}
\label{51}
 \end{equation} 
 Dividing \eqref{51} by $\frac{t}{2}$ yields:
 \begin{equation} 
\frac{2q(t)}{t} e^{i\theta_q}-v_{A/B}e^{i\theta_v}=v_p(t)ie^{i\theta(t)}-\frac{2r}{t}e^{i\theta(t)}
\label{divided}
 \end{equation} 
Let's denote:   $v_q(t)=\frac{2q(t)}{t}$, $v_r(t)=\frac{2r(t)}{t}$, and substitute in \eqref{divided} to yield:
\begin{equation} 
v_q(t) e^{i\theta_q}-v_{A/B}e^{i\theta_v}=v_p(t)ie^{i\theta(t)}-v_r(t) e^{i\theta(t)}.
\label{alfa1}
 \end{equation} 

Solving \eqref{alfa1} for $v_p$:
map t to ct

\begin{equation} 
v_p(ct) =\pm \sqrt{ v_q^2 +v_{A/B}^2 +v_qv_{A/B}\cos(\theta_q-\theta_v)  -v_r^2 }. 
\label{vp}
\end{equation} 
Having solved for $v_p$ \eqref{vp}, we can now solve for $\theta$ at which $v_p$ is tangent to $\partial B$.  

Subtracting the two vectors on the left-hand side of \eqref{alfa1} yields:  
\begin{equation} 
v_\alpha(t) e^{i\alpha(t)}=v_q(t) e^{i\theta_q}-v_{A/B}e^{i\theta_v}
\end{equation} 
Substituting back in \eqref{alfa1}:
\begin{equation} 
v_\alpha(t) e^{i\alpha(t)}=(\pm iv_p(t)-v_r(t))e^{i\theta(t)}
\label{vqa}
\end{equation} 
Dividing both sides by $e^{i\theta(t)}$ yields
\begin{equation} 
v_\alpha(t) e^{i(\alpha(t)-\theta(t))}=\pm iv_p(t)-v_r(t) 
\label{vqa2}
\end{equation} 
Separating \eqref{vqa2} to real and imaginary terms: 

\begin{equation} 
v_\alpha(t)\sin(\alpha(t)-\theta(t))=\pm v_p(t)
\label{sin}
\end{equation} 
\begin{equation} 
v_\alpha(t)\cos(\alpha(t)-\theta(t))=-v_r 
\label{cos}
\end{equation} 
Dividing \eqref{sin} by \eqref{cos}:
\begin{equation} 
\tan(\alpha(t)-\theta(t))= \frac{\pm v_p(t)}{-v_r(t)}
\end{equation} 
Solving for $\theta(t)$:
\begin{eqnarray}  
\theta_R(t) = \alpha(t)+ \tan^{-1} (\frac{ v_p(t)}{v_r(t)})\\
\theta_L(t) = \alpha(t) -\tan^{-1} (\frac{ v_p(t)}{v_r(t)}).
  \end{eqnarray} 
We can now express the grazing accelerations $a_A$ as  functions of $\theta(t)$ and define the boundary of $NAO$:

\begin{eqnarray} 
a_{AR}(t)&=&(v_p(t)e^{i(\theta_R(t)+\pi/2)}-v_Ae^{i\theta_v})/t
\\
a_{AL}(t)&=&(v_p(t)e^{i(\theta_L(t)+ \pi/2)}-v_Ae^{i\theta_v})/t
 \end{eqnarray}
\vspace{5pt}
\noindent \textbf{Definition 2:} Boundary of $NAO$
\begin{equation} 
NAO_R=\{a_{AR}(t)\}\; ;t\in(t_0,t_h]
\end{equation} 
\begin{equation} 
NAO_L=\{a_{AL}(t)\}\; ;t\in(t_0,t_h]
\end{equation} 

\section{Examples}

\begin{figure}%[h]
\vspace{.1cm}
\centerline{\resizebox{8cm}{!}{\includegraphics{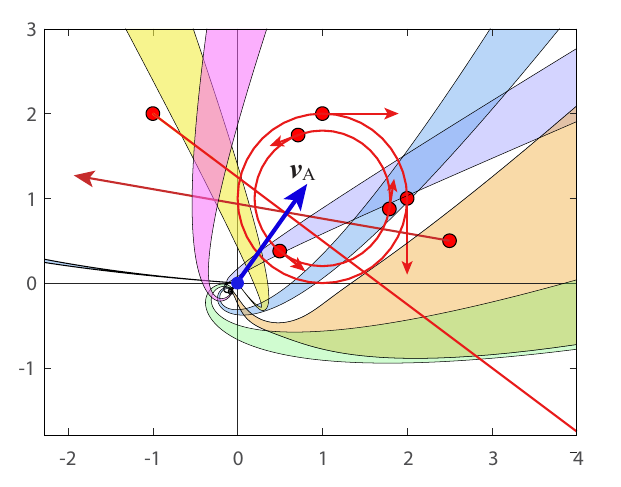}}}
   \caption{The  NAO's of obstacles moving along circular and straight line trajectories for robot A that is located at the origin of the coordinate system and having an initial velocity $v_A$.   }
   \label{naotraj}
 \end{figure}

Fig. \ref{naotraj} shows the NAO's of several obstacles moving along circular and straight line trajectories for robot A that is moving at an initial velocity $v_A$.  Selecting a constant acceleration $a_A$ that points to any of the empty spaces that are not in any NAO guarantees a safe crossing of all obstacles.       

\begin{figure}%[h]
\vspace{.1cm}
\centerline{\resizebox{8cm}{!}{\includegraphics{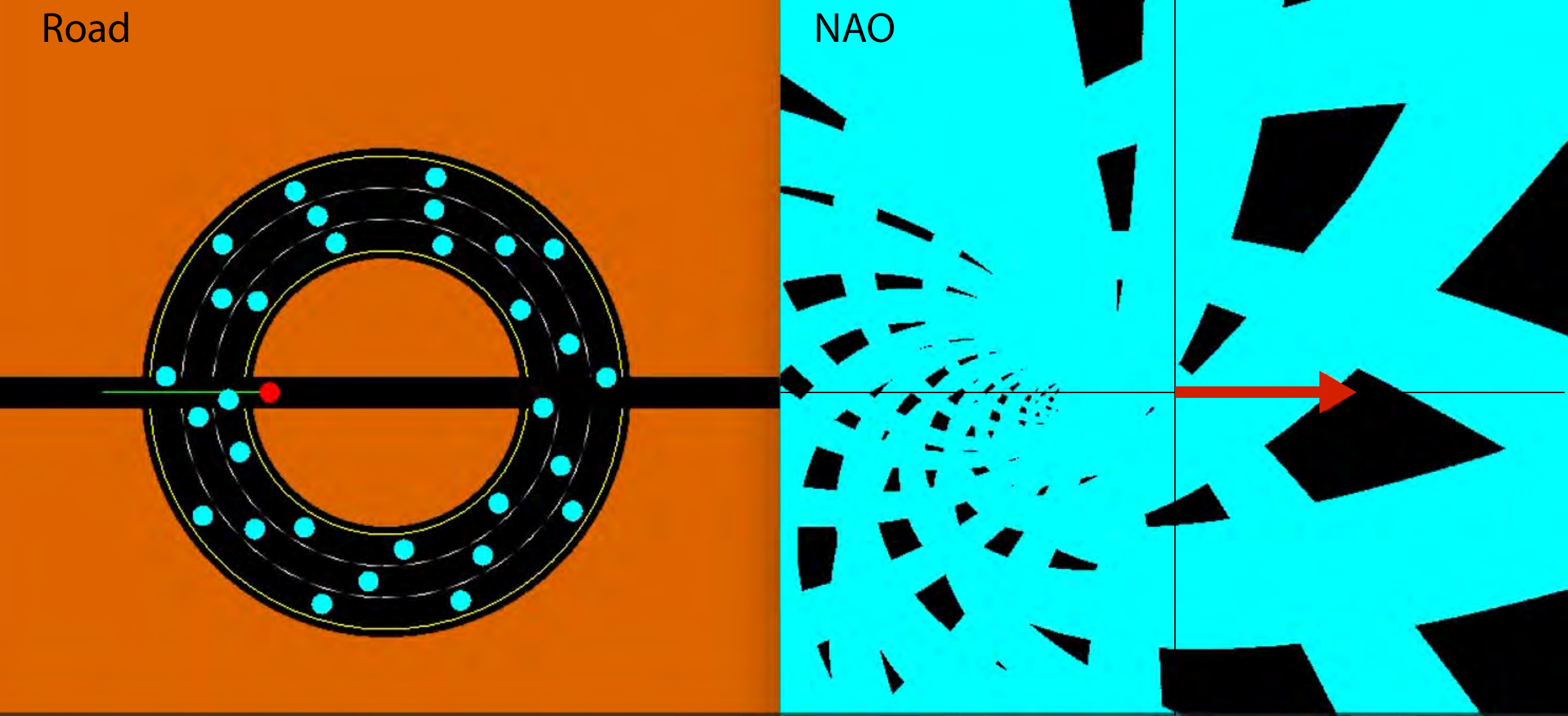}}}
   \caption{Crossing a busy roundabout using NAO.  }
   \label{roundnao}
 \end{figure}

Fig. \ref{roundnao} shows a vehicle (in red) crossing a busy traffic circle with  30 vehicles that are moving in three circular lanes.  The vehicle is crossing the circle at a constant acceleration, selected on the NAO map shown on the right. The selected acceleration is shown as a red arrow on the NAO map.  The vehicle crossed all obstacles with no collision.  Attempting to do the same with AO resulted in many collisions between the crossing and the circling vehicles.  

\begin{figure}%[h]
\vspace{.1cm}
\centerline{\resizebox{8cm}{!}{\includegraphics{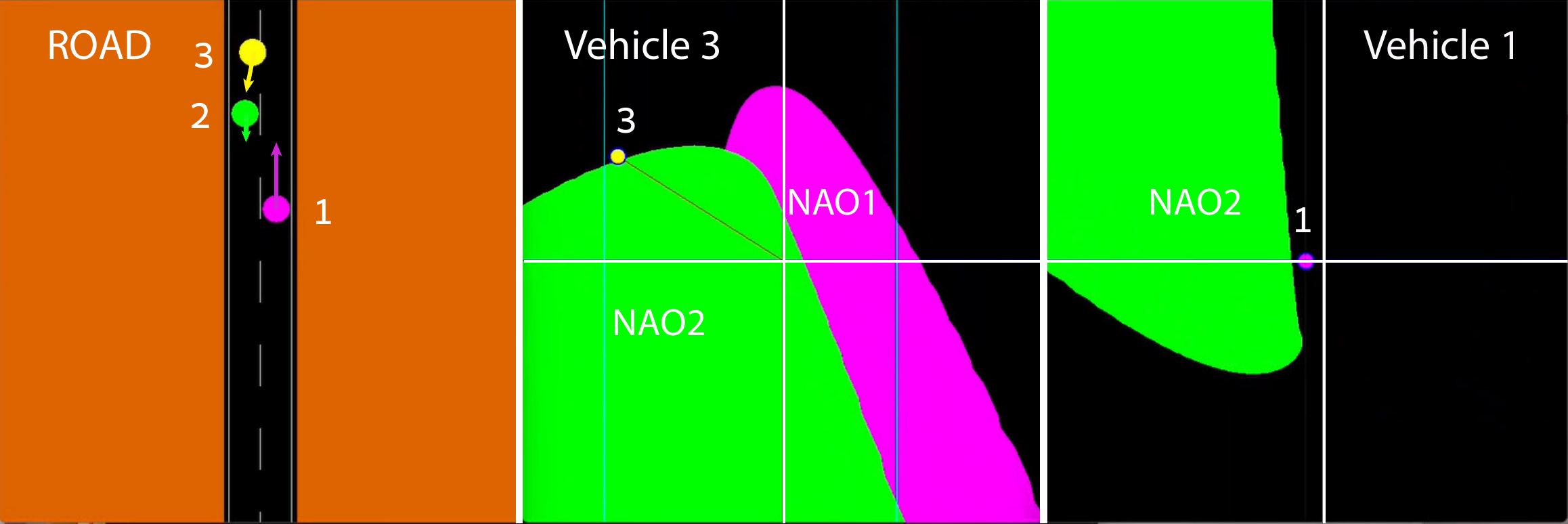}}}
   \caption{Overtaking safely a slow vehicle on a two-lane road, one lane for each direction, using NAO.}
   \label{road1}
 \end{figure}

Fig. \ref{road1} shows three vehicles marked 1,2,3, on a two-lane road.   Vehicle (in tellow) attempts to overtake a slow moving vehicle 4 (in green).  However, NAO1 of vehicle 1 (in purple) prevents it from accelerating to the right.  Vehicle 3 is shown as a yellow dot in its NAO map of vehicles 1 and 2. It moves back to the left and slows down behind vehicle 2--its acceleration is negative and pointing to the left.  Vehicle 1, marked by a purple dot on its NAO map, moves at a constant desired speed with zero forward acceleration.

These examples demonstrate the usefulness of the NAo in negotiating complex dynamic environments,  and in locally selecting dynamically feasible collision-avoiding accelerations. The resulting vehicle motions are smooth, resembling the behavior of careful experienced drivers.   
     
\section{Conclusions}
 
The concept of Velocity Obstacles was extended to Acceleration Obstacles $AO$ and Nonlinear Acceleration Obstacles $NAO$ to allow the maneuvering vehicle to use its acceleration to avoid collisions in complex dynamic environments.   
The $AO$ and $NAO$ were defined by their exact boundaries, derived analytically for efficient computation of the avoiding accelerations.  Using nonlinear acceleration obstacles allows for
more efficient avoidance maneuvers (fewer adjustments) than the linear acceleration obstacle for the case of obstacles moving along general trajectories.  The result is safer avoidance maneuvers in complex situations, as was demonstrated in several challenging scenarios. As we study the properties of the newly developed NAO, more challenging test cases will be presented in the near future.    
 
\bibliographystyle{plain}

\bibliography{main}

\end{document}